# Modeling Adaptive Platoon and Reservation Based Autonomous Intersection Control: A Deep Reinforcement Learning Approach

Duowei LI[1,2], Jianping WU[1], Feng ZHU*[,2], Tianyi CHEN[2], and Yiik Diew WONG[2]

[1] Department of Civil Engineering, Tsinghua University, China

[2] School of Civil and Environmental Engineering, Nanyang Technological University, Singapore

## SHORT SUMMARY

As a strategy to reduce travel delay and enhance energy efficiency, platooning of connected and autonomous vehicles (CAVs) at non-signalized intersections has become increasingly popular in academia. However, few studies have attempted to model the relation between the optimal platoon size and the traffic conditions around the intersection. To this end, this study proposes an adaptive platoon based autonomous intersection control model powered by deep reinforcement learning (DRL) technique. The model framework has following two levels: the first level adopts a First Come First Serve (FCFS) reservation based policy integrated with a nonconflicting lane selection mechanism to determine vehicles' passing priority; and the second level applies a deep Q-network algorithm to identify the optimal platoon size based on the real-time traffic condition of an intersection. When being tested on a traffic micro-simulator, our proposed model exhibits superior performances on travel efficiency and fuel conservation as compared to the state-of-the-art methods.

## 1. INTRODUCTION

Intersection management is essential for operation efficiency and safety of a traffic system. With the emergence of new technologies, intersection control is expected to usher in new breakthroughs. On one hand, the large-scale deployment of roadside sensors can provide real time traffic data for adaptive traffic signal control. For example, signal lights can dynamically adjust phase cycle and even sequence according to the data collected from each intersection and optimize the signal control over a large-scale road network (Li et al., 2020a). On the other hand, with the rise of connected and autonomous vehicles (CAVs), many researchers have focused on managing non-signalized intersections by involving CAVs. Dresner and Stone (2004) developed the Autonomous Intersection Management (AIM) protocol which can coordinate multiple CAVs to pass through an intersection simultaneously based on First Come First Serve (FCFS) policy and significantly reduces traffic delay as compared to signalized intersection. Taking the advantages of CAVs over human driving vehicles (HVs), such as higher predictability, shorter distance headway, quicker response, etc., platooning CAVs can improve intersection management by grouping the vehicles with similar attributes into an indivisible queue. Platooning has the following characteristics: (1) The communication burden can be significantly reduced as the intersection manager (IM) only communicates with the platoon leader instead of with each CAV (Kumaravel et al., 2021); (2) The road capacity and operational efficiency of an intersection can be increased significantly due to close and insert-forbidden vehicle-following behaviors (Lioris et al., 2017); and (3) Platooning can reduce energy consumption by lowering the frequency of acceleration and deceleration (Li et al. 2020b).

The focuses of previous research in platooning CAVs at the non-signalized intersection can be classified into two categories: centralized reservation based control and decentralized optimal control. Jin et al. (2013) proposed a two-level centralized reservation based strategy that can reduce the communication load by 90% while ensuring efficiency. The grouping algorithm divides the CAVs into multiple platoons composed of the leader vehicle agent (LVA) and the follower vehicle agents (FVAs). The reservation process only involves the information interacting between IM and LVA, while LVA further plans the trajectory for the platoon based on the reservation information. In the decentralized optimal control framework proposed by Kumaravel et al. (2021), only V2V communication is involved where each platoon leader communicates with other platoon leaders and a coordinator derives the optimal schedule to cross the intersection. Previous studies have demonstrated the reliability and efficiency of capsulating platoons into both centralized and decentralized intersection controls. However, although platoon size is of significance to the traffic efficiency at intersection, few of those studies have attempted to model the relation between the optimal platoon size and the traffic conditions around the intersection. Specifically, most previous studies failed to judge whether a CAV should join or leave a platoon and determine a suitable size (or length) for a platoon by comprehensively considering the traffic conditions (e.g., traffic density, vehicles' movement, etc.). Instead, they either force the CAVs within V2V range to join the platoon or to form a fixed length platoon.

To bridge the abovementioned research gap in the previous studies, this study proposes an adaptive platoon based autonomous intersection control model, which is powered by a deep reinforcement learning (DRL) technique. Besides, this proposed model also makes the following contributions to better resolve the research gap:
(1) The model ensures safety, efficiency and fairness when platooning CAVs and releasing the platoons to pass through the intersection.
(2) An algorithm is implemented to achieve adaptive and dynamic platooning, where the optimal platoon size is identified based on the real-time traffic conditions.
(3) DRL is applied in the platoon control at intersection and represents the problem-related elements as the DRL components.
(4) A phase selection mechanism is embedded in the model to allow CAVs to make nonconflicting movement at intersection zone simultaneously.

The rest of this paper is organized as follows: Section 2 introduces the general framework as well as the components of the proposed model, Section 3 presents the experiment of applying the proposed model on a simulator and discusses the results, and Section 4 covers the conclusion of this study.

## 2. METHODOLOGY

*General framework*

As shown in Figure 1, the proposed model has a two-level framework. To simplify the description, this study divides an intersection area into two zones, namely the square intersection zone ($length = S$) where conflicts may happen, and the circular control zone ($radius = L$) where V2I and V2V communication occurs, as illustrated in Figure 2. The first level refers to reservation based lane identification level, where the target lane with the highest priority is identified mainly based on a FCFS policy. A nonconflicting lane selection mechanism is embedded in this level to ensure the safety and efficiency by allowing CAVs with nonconflicting movement to be in the intersection zone simultaneously. The second level mainly refers to DRL based platoon size determination level. Once the optimal platoon size of the target lane is determined by the DRL agent, the IM will allow the platoon in the target lane as well as CAVs in the selected nonconflicting lanes to pass through without any conflicts. The following two sub-sections describes the above-mentioned two levels in details respectively.

The framework is constructed with following assumptions:
(1) Communication delay and errors between IM, LVAs and FVAs are not considered.
(2) CAVs within a platoon move at the same speed and maintain a constant desired distance headway.
(3) Lane-changing behaviors are not allowed in the control zone, which means all CAVs should proactively move to a desired lane before sending a request.
(4) CAVs should send a request at the moment when entering the control zone.
(5) CAVs should move along a pre-defined trajectory in the intersection zone as shown in Figure 3.

*First level: Reservation based lane identification level*

In this study, LVAs, FVAs and IM coordinate their behaviors by using the messages as listed in Table 1. The four categories of interactions, namely CAV to IM, IM to LVA, LVA to FVA, and LVA to IM happen sequentially. The timestamps of the requests sent from CAVs to IM are recorded and sorted. The FCFS policy ensures the lane with the earliest request to have the priority to form a platoon and pass through the intersection zone. Furthermore, in order to increase efficiency, CAVs in optimal nonconflicting lanes can also be released at the same time. Take the target lane (North-South Straight Lane) in Figure 4(a) as an example, there are three corresponding combinations of optional nonconflicting lanes as shown in Figures 4(b)-4(d). The lane that satisfies Equation (1) is chosen as the optimal nonconflicting lane $l_{optimal}$:

$$l_{optimal} = arg\max_{l \in L} \sum_{v \in V_l} f_{waited}(v) \qquad (1)$$

where $f_{waited}$ is the accumulative waited time, $L$ is the set of nonconflicting lanes, and $V_l$ denotes the set of CAVs traveling on a nonconflicting lane $l$.

**Table 1: Message categories in communication protocol.**

| Message category | CAV to IM | IM to LVA | LVA to FVA | LVA to IM |
| --- | --- | --- | --- | --- |
| Content | Vehicle ID<br>Current position<br>Current speed<br>Acceleration limit<br>Turning demand | Departure time<br>Platoon size<br>Desired speed | Join or not<br>Head spacing<br>Follow speed | Passed or not |

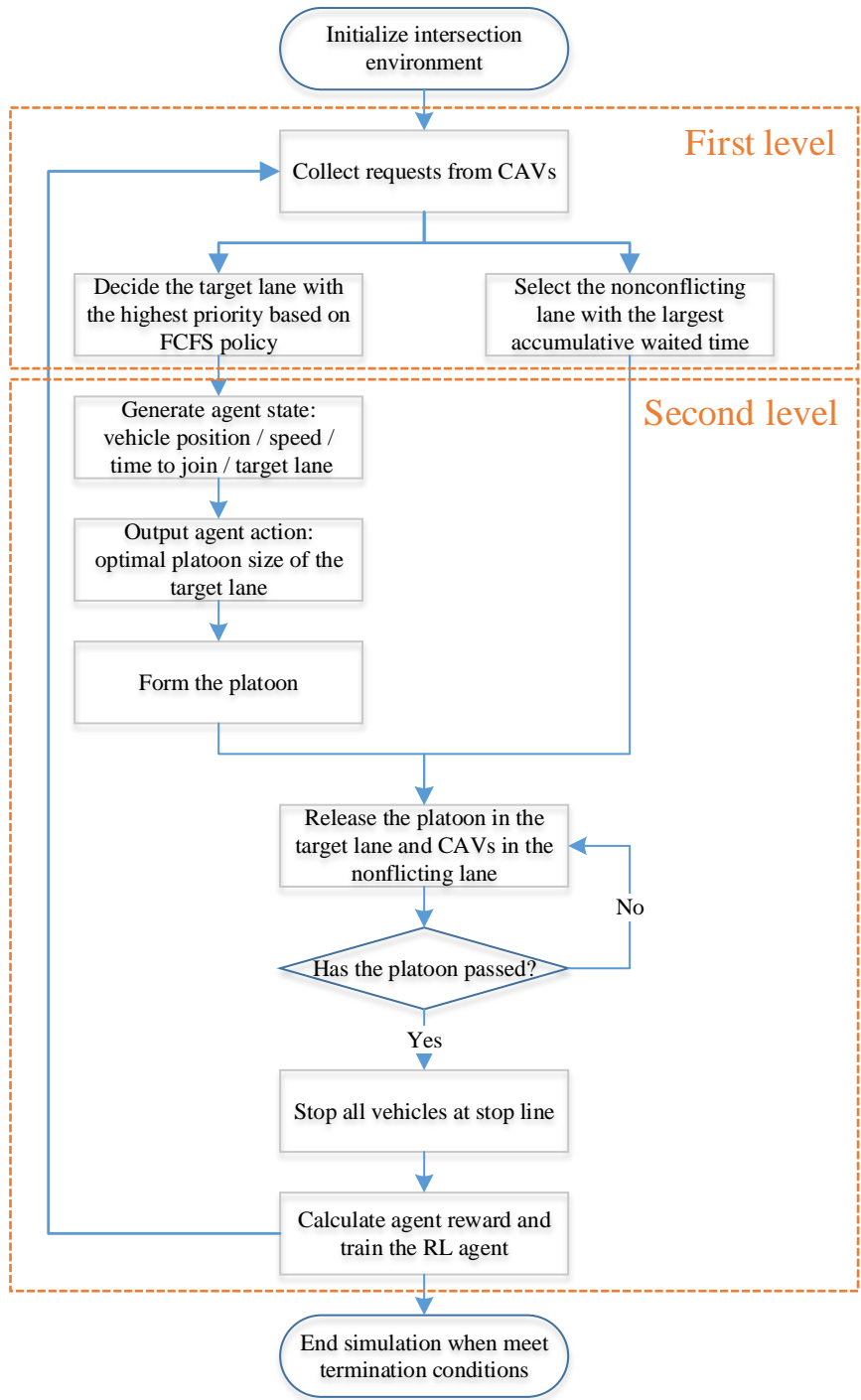

**Figure 1: Flowchart of two-level model framework.**

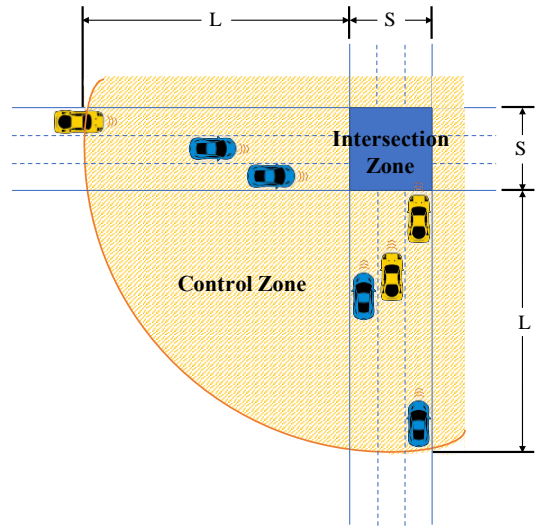

**Figure 2: Illustration of intersection area.**

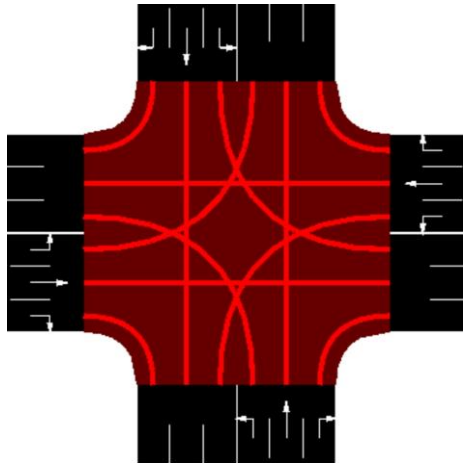

**Figure 3: Pre-defined trajectories in intersection zone. Each turning trajectory is a quarter arc.**

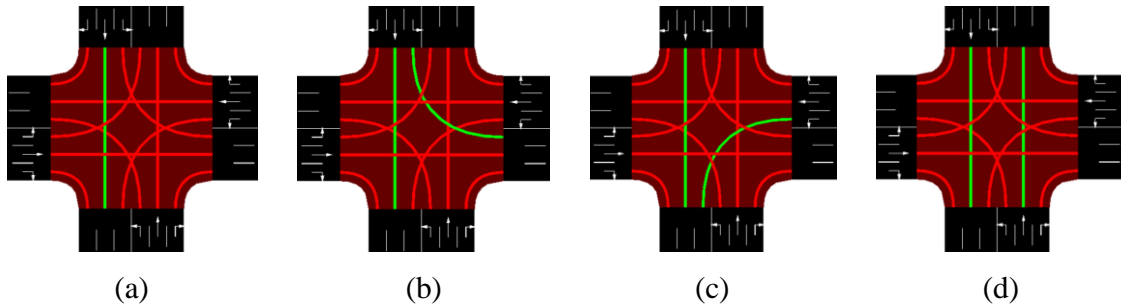

(a)          (b)          (c)          (d)

**Figure 4: Examples of nonconflicting lane selection mechanism. (a): Target lane; and (b)-(d): Combinations of optional nonconflicting lanes.**

*Second level: DRL based platoon size determination level*

Trial-and-error search and delayed reward are the two most important distinguishing features that make RL suitable for platoon control. RL can precisely represent the elements associated with the problem: agent (IM), environment (traffic condition), and actions (platoon size). Deep Q-network (DQN), is applied to determine the optimal platoon size in this study, as it synthesizes the benefits of both RL and convolutional neural networks (CNNs), which enables it to sufficiently extract the spatial characteristics around an intersection (Li et al. 2021). The pseudocode of training DQN is given in Algorithm 1.

---
**Algorithm 1** DQN with experience replay
---
1. **Definition**
2.    $D \coloneqq$ replay memory pool
3.    $N \coloneqq$ maximum number of experiences in $D$
4.    $Q \coloneqq$ action-value function in Eval_net
5.    $\hat{Q} \coloneqq$ action-value function in Target_net
6.    $M \coloneqq$ maximum number of episode
7.    $T \coloneqq$ maximum number of iteration in each episode
8. **Initialization**
9.    $D \leftarrow$ Initial replay memory to capacity $N$
10.   $Q \leftarrow$ Initial evaluate action-value function with random weights $\theta$
11.   $\hat{Q} \leftarrow$ Initial target action-value function with random weights $\theta^- = \theta$
12. **For** episode $= 1, M$ **do**
13.     Observe n steps before decision-making
14.     Initialize environment state $s_1$
15.    **For** $t = 1, T$ **do**
16.       With probability $\varepsilon$ select a random action $a_t$
17.       Otherwise select $a_t = argmax_a Q(s_t, a; \theta)$
18.       Execute action $a_t$ in SUMO and observe reward $r_t$ and environment state $s_{t+1}$
19.       Store experience $e_t = (s_t, a_t, r_t, s_{t+1})$ in $D$
20.       Sample random *batch_size* experiences $e_j = (s_j, a_j, r_j, s_{j+1})$ from $D$
21.       Set $y_j = \begin{cases} r_j, & \text{if episode terminates at step j+1} \\ r_j + \gamma max_{a'} \hat{Q}(s_{j+1}, a'; \theta^-), & \text{otherwise} \end{cases}$
22.       Updating network parameters $\theta$ by perform a gradient decent step on $(y_j - Q(s_j, a_j; \theta))^2$
23.       Every $C$ steps reset $\hat{Q} = Q$
24.       Set $s_t = s_{t+1}$
25.    **End for**
26. **End for**
---

The three most essential parts of the DQN agent are state space *S*, action space *A*, and reward *R*. In this study, in order to ensure that the environment is accurately represented and also comprehensively captured by CNNs, the environmental state is processed as four matrixes in the model: CAV locations, CAV speed, time to join the platoon, and a map of the target lane. Figure 5(a) illustrates a part of the control zone, where the red line on the rightmost represents the stop line, and the dotted lines grid the interface into squares with side length of each square equal to lane width. Figure 5(b) represents the occupied state of each square and the value of 1 means more than half of the square area is occupied by a CAV. Figure 5(c) exhibits the speed (m/s) of the CAV occupying the corresponding square. The value in Figure 5(d) refers to the time spent by the corresponding CAV to join a platoon, $t_{TTJ}$, which can be calculated from:

$$v_c^i t_{TTJ}^i + \frac{1}{2}a_{max}^i {t_{TTJ}^i}^2 + \sum_{j=1}^{i-1} l_c^j + (i-1)d_h = v_c^0 t_{TTJ}^i + d_{1,i} \quad (2)$$

where $v_c^i$, $a_{max}^i$, $l_c^i$ and $t_{TTJ}^i$ respectively indicate the current speed, maximum acceleration, vehicle length and time-to-join of the $i^{th}$ CAV in a lane, $d_{1,i}$ denotes the distance between the front of the first CAV and the front of the $i^{th}$ CAV, and $d_h$ is the desired distance headway. Note that all the above-mentioned environmental state can be acquired by the requests sent from CAV to the IM in Table 1.

The action space $A = \{1,2,\dots,N\}$ is a set of the feasible platoon sizes of the target lane, which shall meet the following constraints in Equation (3) and Equation (4) to ensure that the platoon length is smaller than the control zone length.

$$N = \underset{n}{arg\max}\, L_n := \{n \mid L_n \leq L\} \quad (3)$$

$$L_n = \sum_{i=1}^{n} l_c^i + (n-1)d_h \quad (4)$$

At each timestamp, all the CAVs at an intersection are iterated and their waited time $W_i$ (spent before the stop line) is recorded. The reward $R$ is calculated by Equation (5) to make it inversely proportional to the average waiting time of each CAV, upon which the optimal platoon size with maximum reward can be identified.

$$R_i = c - c\left(\frac{W_i}{W_m}\right)^2 \quad (5)$$

Notice that $R_i$ will become negative when $W_i$ reaches a threshold value $W_m$, which suggests that the CAV has waited for too long and should be released. The constant $c$ is a parameter to control the upper bound of $R_i$.

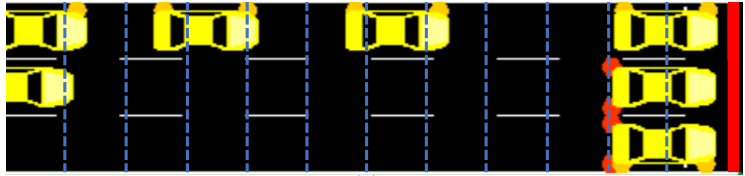

(a)

| 1 | 1 | 1 | 1 | 0 | 0 | 1 | 1 | 0 | 0 | 1 | 1 |
|---|---|---|---|---|---|---|---|---|---|---|---|
| 1 | 0 | 0 | 0 | 0 | 0 | 0 | 0 | 0 | 0 | 1 | 1 |
| 0 | 0 | 0 | 0 | 0 | 0 | 0 | 0 | 0 | 0 | 1 | 1 |

(b)

| 17.9 | 17.9 | 14.0 | 14.0 | 0 | 0 | 12.5 | 12.5 | 0 | 0 | 5.2 | 5.2 |
|---|---|---|---|---|---|---|---|---|---|---|---|
| 20.1 | 0 | 0 | 0 | 0 | 0 | 0 | 0 | 0 | 0 | 0 | 0 |
| 0 | 0 | 0 | 0 | 0 | 0 | 0 | 0 | 0 | 0 | 0 | 0 |

(c)

| 0.62 | 0.62 | 0.60 | 0.60 | 0 | 0 | 0.46 | 0.46 | 0 | 0 | 0 | 0 |
|---|---|---|---|---|---|---|---|---|---|---|---|
| 0.75 | 0 | 0 | 0 | 0 | 0 | 0 | 0 | 0 | 0 | 0 | 0 |
| 0 | 0 | 0 | 0 | 0 | 0 | 0 | 0 | 0 | 0 | 0 | 0 |

(d)

**Figure 5: Example of processing traffic environment into state maps. (a) Simulation interface. (b) CAV location matrix. (c) CAV speed matrix. (d)Time to join the platoon matrix.**

## 3. RESULTS AND DISCUSSION

*Experiment design and parameter setting*

The proposed model is validated and verified on the traffic micro-simulator "Simulation of Urban Mobility" (SUMO). The simulation network environment is shown in Figure 3, which is a standard intersection with each direction consisting of a left-only lane, a straight-only lane, and a right-only lane. The parameter settings of the model and the simulated environment are listed in Table 2. As shown in Table 3, we adopt unevenly distributed traffic flows to examine the model performance in a relatively complex environment.

*Method comparison*

Four state-of-the-art methods are employed for the comparison with our proposed model:
(1) **Webster fixed signal timing control.** Four traffic signal phases are periodically changed in a round-robin manner and the phase duration is designed using Webster method (Webster, 1958).
(2) **Non-platooning reservation based method**. FCFS policy is adopted for the reservations of individual CAVs (Dresner and Stone, 2004).
(3) **Fixed length platoon based method**. The platoons are defined with fixed sizes of 3, 6, 9 and 12, while the other settings are same with those of our proposed model.
(4) **Basic adaptive platoon method.** The nonconflicting lane are randomly selected instead of using nonconflicting lanes selection mechanism, while the other settings are same with those of our proposed model.

**Table 2: Parameter settings.**

| Parameters | Values | Explanations |
|---|---|---|
| $l_c$ | 5.0 (m) | Length of CAVs |
| $w_c$ | 1.8 (m) | Width of CAVs |
| $l_{lane}$ | 2.5 (m) | Width of traffic lanes |
| $S$ | 15.0 (m) | Length of the intersection zone |
| $L$ | 200.0 (m) | Length of the control zone |
| $a_{max}$ | ±5 (m/s$^2$) | Maximum acceleration and deceleration of CAVs |
| $v_{max}$ | 20 (m/s) | Upper limit speed in the network |
| $d_h$ | 1.0 m | Desired distance headway in the platoon |
| $d_{\hat{h}}$ | 1.5 m | Minimum distance headway outside the platoon |
| $T$ | 3600 (s) | Simulation duration in each episode |
| $M$ | 100 | Maximum number of training episodes |
| $\varepsilon$ | 0.1 | ε_greedy strategy：10% Exploration and 90% Exploitation |
| *replay_memory size* | 1000 | Maximum size of the memory pool |
| *batch_size* | 32 | Size of memory extracted from the pool for learning each time |
| *observe step* | 100 | Number of steps to observe before training process |
| $c$ | 0.15 | The upper bound of reward |
| $W_m$ | 60 | Threshold value of $W_i$ when reward becomes negative |

**Table 3: Configurations of simulated traffic environment.**

| Direction   | Flow rate (veh/h) | Direction   | Flow rate (veh/h) |
|-------------|-------------------|-------------|-------------------|
| North-South | 500               | East-North  | 400               |
| North-East  | 300               | East-South  | 500               |
| North-West  | 200               | East-West   | 700               |
| South-North | 400               | West-North  | 300               |
| South-East  | 400               | West-South  | 200               |
| South-West  | 200               | West-East   | 500               |

*Performance analysis*

The performances on average travel time and fuel consumption (measured according to Handbook on Emission Factors for Road Transport (HBEFA)) of each method for comparison are reported in Table 4. It is found that the proposed model outperforms the other methods on those two aspects. Traditional fixed signal timing is unable to take full advantage of CAVs and make efficient control. Non-platooning reservation based method performs even worse due to the reduction of throughput by serving limited number of CAVs simultaneously. The performances of the basic adaptive platoon method are slightly less satisfactory as compared to our proposed method. Such differences in the performances can manifest the necessity of adopting the nonconflicting lane selection mechanism to a certain extent. The fixed length platoon based method attains its optimal performances when the platoon length is six. This is to a certain degree in line with a finding relevant to the distribution of platoon sizes in our proposed model, which is illustrated in Figure 6. It is found that the platoon sizes are not evenly distributed and the three most frequent platoon sizes are 7, 6 and 8, respectively. The above comparison results manifest the superiority of our proposed model on identifying optimal platoon size, which achieves satisfactory performances on both travel efficiency and energy conservation.

**Table 4: Reports of method comparison.**

| Methods                                     |     | Average travel time (s) | Fuel consumption (mL/veh) |
|---------------------------------------------|-----|-------------------------|---------------------------|
| Webster method                              |     | 110.87                  | 126.03                    |
| Non-platooning reservation based method     |     | 134.16                  | 116.74                    |
| Fixed platoon based method with size of     | 3   | 108.18                  | 100.61                    |
|                                             | 6   | 74.42                   | 95.89                     |
|                                             | 9   | 84.85                   | 104.22                    |
|                                             | 12  | 104.86                  | 113.64                    |
| Basic adaptive platoon method               |     | 72.42                   | 90.91                     |
| Proposed model                              |     | 69.87                   | 89.29                     |

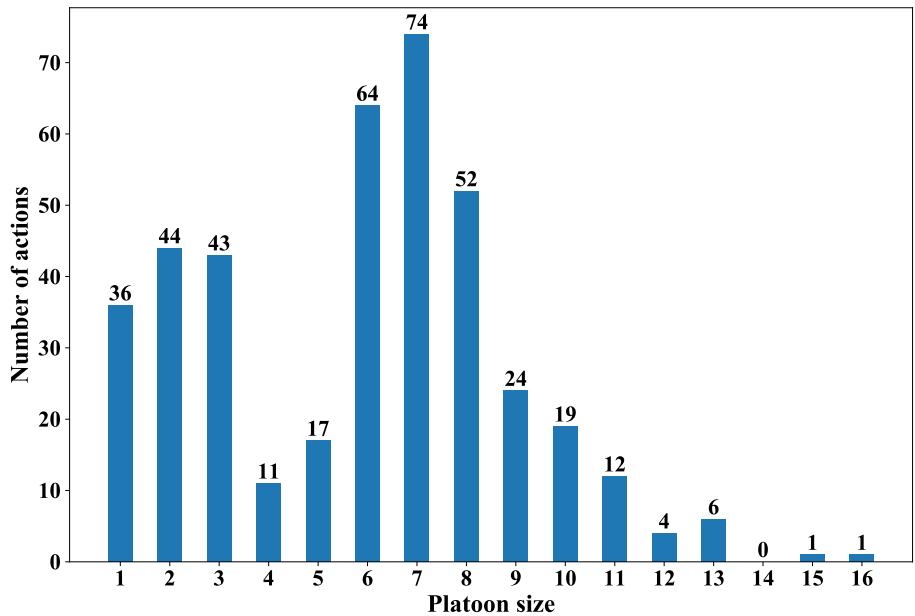

Figure 6: Distribution of platoon size.

## 4. CONCLUSIONS

In this study, we propose an autonomous intersection control model which can adaptively platoon CAVs based on the multiple characteristics of traffic conditions. The two-level framework of the model involves reservation based lane identification and DRL based platoon size determination, which ensures the platoon in the target lane as well as CAVs in the selected nonconflicting lanes to pass through the intersection efficiently and safely. The model is validated and verified in SUMO under unevenly distributed traffic flows. The comparison with the other state-of-the-art methods demonstrates the superior performances of the proposed model in reducing travel time and fuel consumption. In the future, multiple intersection scenarios and traffic conditions can be further incorporated into the construction and verification of the model.

## ACKNOWLEDGMENT

The first author would like to acknowledge the State Scholarship Fund provided by the China Scholarship Council that supports her studies in Nanyang Technological University Singapore.

## REFERENCES

Dresner K., Stone P. 2004. Multiagent traffic management: a reservation-based intersection control mechanism. In Proceedings of the 3$^{rd}$ International Joint Conference on Autonomous Agents and Multiagent Systems, 19-23 July, New York, USA, pp. 530-537.

Jin Q., Wu G., Boriboonsomsin K., Barth M. 2013. Platoon-based multi-agent intersection management for connected vehicle. In Proceedings of the 16th International IEEE Conference on Intelligent Transportation Systems, 6-9 October, The Hague, Netherlands, pp. 1462–1467.

Kumaravel SD., Malikopoulos A., Ayyagari R. Optimal Coordination of Platoons of Connected and Automated Vehicles at Signal-Free Intersections. 2021. *IEEE Transactions on Intelligent Vehicles*. pp. 1-1.

Li D., Wu J., Peng D. 2021. Online Traffic Accident Spatial-Temporal Post-Impact Prediction Model on Highways Based on Spiking Neural Networks. *Journal of Advanced Transportation*. Vol.2021.

Li D., Wu J., Xu M., Wang Z., Hu K. 2020a. Adaptive Traffic Signal Control Model on Intersections Based on Deep Reinforcement Learning. *Journal of Advanced Transportation*. Vol.2020.

Li Q., Zhang J., Ji Y., Wu K., Yu T., Wang J. 2020b. Research on Platoon Dynamic Dispatching at Unsignalized Intersections in Intelligent and Connected Transportation Systems. *SAE Technical Paper*. No. 2020-01-5199.

Lioris J., Pedarsani R., Tascikaraoglu FY., Varaiya P. 2017. Platoons of connected vehicles can double throughput in urban roads. *Transportation Research Part C: Emerging Technologies*. Vol.77, pp. 292-305.

Webster FV. 1958. Traffic signal settings. *Road Research Lab Tech Papers.*